\documentclass[11pt]{cambrian}

\usepackage[most]{tcolorbox}
\usepackage{float}
\usepackage{placeins}
\usepackage[capitalize,noabbrev]{cleveref}

\definecolor{mistteal}{RGB}{15,110,86}
\definecolor{mistbrown}{RGB}{139,94,75}
\definecolor{mistgold}{RGB}{176,141,87}
\definecolor{mistblue}{RGB}{91,124,153}
\definecolor{mistpurple}{RGB}{91,78,139}
\definecolor{bluelink}{RGB}{0,113,188}
\definecolor{bestrow}{RGB}{232,243,239}
\hypersetup{colorlinks=true,citecolor=mistteal,linkcolor=black,urlcolor=bluelink}

\newcommand{\mistweb}{\raisebox{-1.5pt}{\includegraphics[height=1.05em]{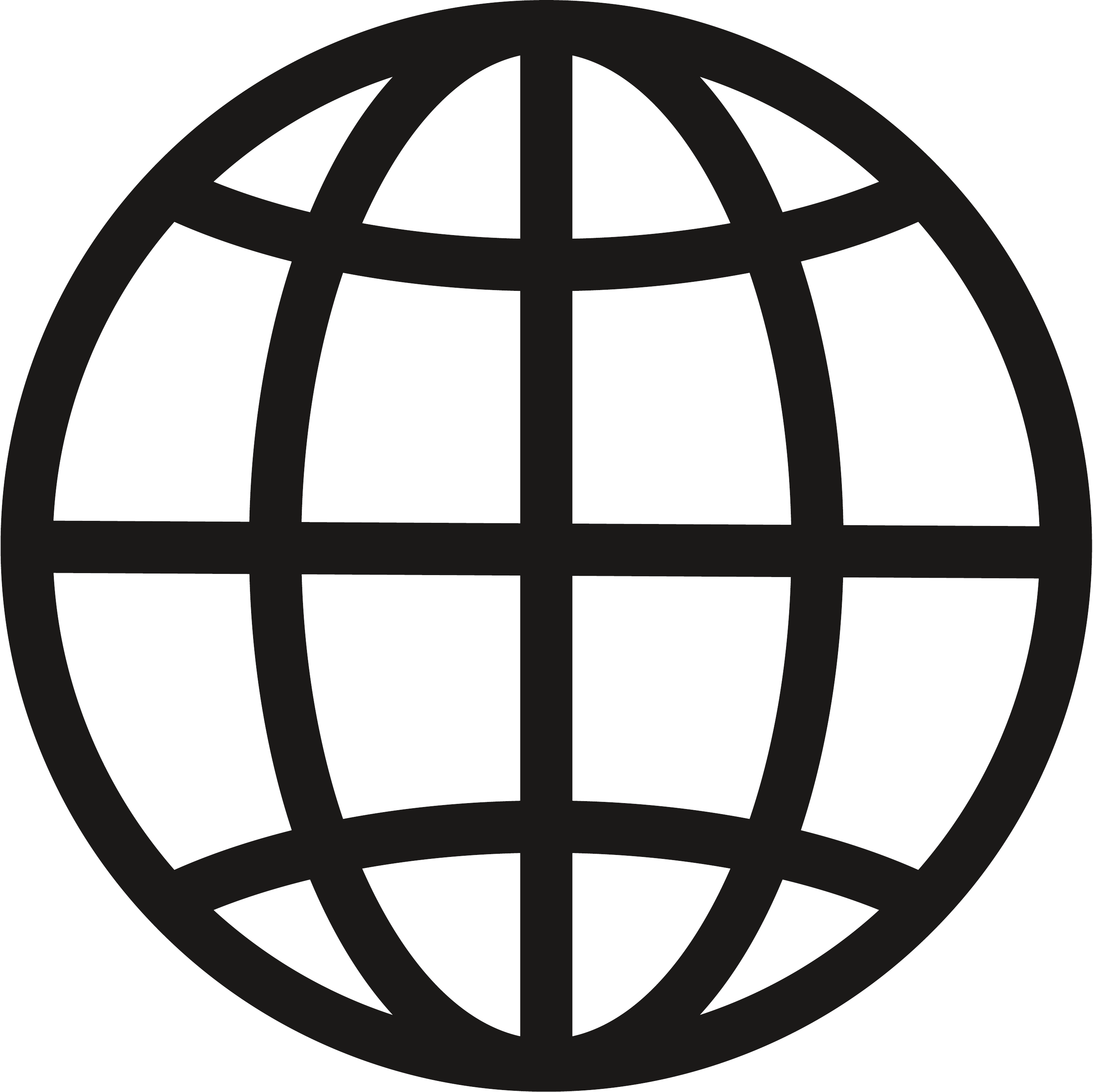}}}
\newcommand{\mistgithub}{\raisebox{-1.5pt}{\includegraphics[height=1.05em]{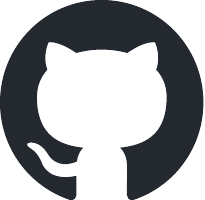}}}

\newcommand{\mistheaderlogos}{%
  \raisebox{-1.2pt}{%
    \includegraphics[height=9mm]{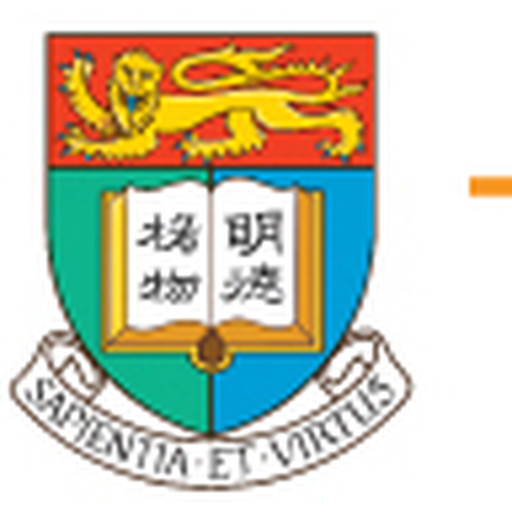}%
    \hspace{4.5pt}%
    \includegraphics[height=9mm]{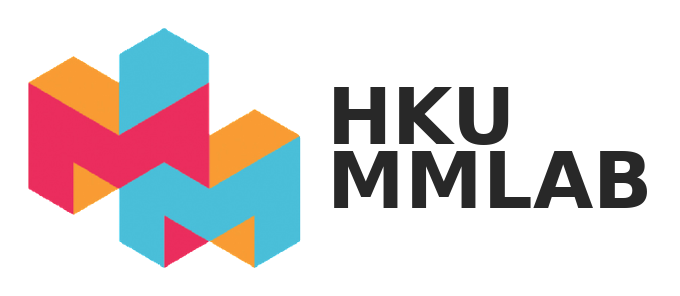}%
    \hspace{4.5pt}%
    \includegraphics[height=9mm]{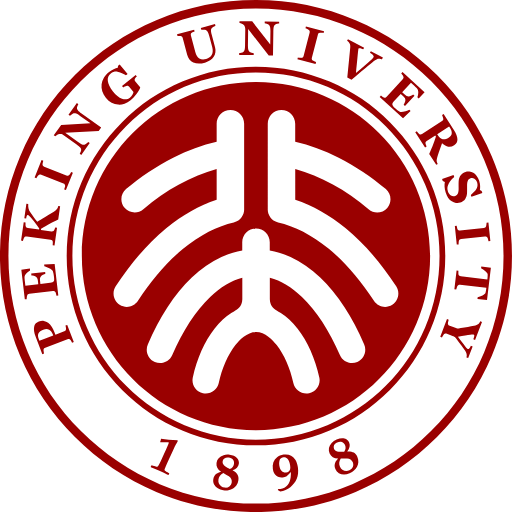}%
    \hspace{4.5pt}%
    \includegraphics[height=9mm]{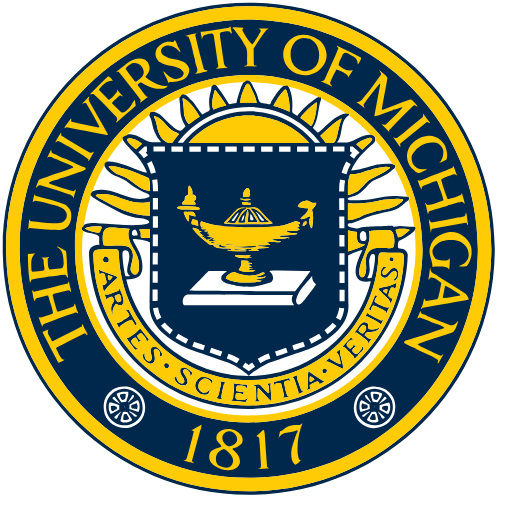}%
  }%
}
\fancypagestyle{firststyle}{%
  \fancyhf{}%
  \fancyhead[R]{\mistheaderlogos}%
  \fancyfoot[L]{%
    \ifdefined\correspondingauthor
    \if\relax\the\correspondingauthor\relax
    \else \footerfont {*\the\correspondingauthor \\} \fi
    \else \fi
  }%
  \renewcommand{\headrulewidth}{1pt}%
  \renewcommand{\footrulewidth}{1pt}%
  \renewcommand{\headrule}{\color{gray}\hrule height \headrulewidth width\headwidth \vskip-\headrulewidth}%
  \renewcommand{\footrule}{\textcolor{gray}{\hrule height \footrulewidth width\headwidth}}%
}

\newcommand{\thinparagraph}[1]{\vspace{0.65em}\noindent\textbf{#1}\enspace}
\newcommand{\finding}[2]{%
  \vspace{0.55em}
  \begin{tcolorbox}[
      enhanced jigsaw,
      colback=mistteal!5,
      colframe=mistteal!80!black,
      boxrule=0.35pt,
      leftrule=1.6pt,
      arc=1.2pt,
      boxsep=0pt,
      left=7pt,
      right=7pt,
      top=4.5pt,
      bottom=4.5pt,
      before skip=4pt,
      after skip=8pt
    ]
    \noindent\textbf{\textit{Finding #1.}}~#2
  \end{tcolorbox}
}

\renewcommand\Affilfont{\normalfont\fontsize{10}{13}\selectfont}
\makeatletter
\renewcommand\AB@affilsepx{\protect\\\protect\Affilfont}
\makeatother

\title{Execution Before Memory:\\Diagnosing Memory Failures in Video World Models}

\author[1,*]{Weiyang Jin}
\author[1,*]{Jiwen Yu}
\author[2]{Sihan Xu}
\author[3]{Yuwei Niu}
\author[1,$\dagger$]{Xihui Liu}
\affil[1]{University of Hong Kong (hku.hk)}
\affil[2]{University of Michigan - Ann Arbor (umich.edu)}
\affil[3]{Peking University (pku.edu.cn)}
\affil[*]{Equal contribution}
\affil[$\dagger$]{Corresponding author}

\begin{abstract}
A memory test for a video world model should say whether the scene
was set, whether the asked change was carried out, and whether the
last frame shows the result. Current tests fold these into one
number, so a low score is easy to read as forgetting.
When the change is not carried out, the hidden state transition
cannot appear, so the test is not really a test of memory.
We introduce MIST, an evaluation that separates the change from the
memory it should leave behind.
The model first sets the scene while the change stays hidden, and
only later is asked to show it, with leakage before that point kept
under control. This separation locates the bottleneck: the model sets
the scene but does not carry out the instructed change.
The same gap appears in general memory evaluations.
On that basis we propose M3-Agent (Make the Missing Move), which recursively revises the details of the input
instruction so that the model successfully carries out the
requested instruction. Relative to the baseline, it more than
triples the MIST reveal score. On the
benchmarks, WRBench's score $T$ rises by 26.1\% and MBench-T's score
$U$ by 52.7\%. The method meets the problem from the execution side:
it shows what the frozen model can do once the move is stated, and it
sets a stricter baseline.
\end{abstract}

\keywords{video world models, memory evaluation, delayed reveal, test-time revision}

\begin{document}
\maketitle
\pagestyle{fancy}

\begin{center}
  \vspace{0.25em}
  \renewcommand{\arraystretch}{1.35}
  \setlength{\tabcolsep}{0pt}
  \begin{tabular}{r@{\hspace{0.5em}}l@{\hspace{0.9em}}l}
    \mistweb{} & \textbf{Website} &
      \href{https://hku-mmlab.github.io/MIST/}{\texttt{hku-mmlab.github.io/MIST}}\\
    \mistgithub{} & \textbf{Code} &
      \href{https://github.com/HKU-MMLab/MIST}{\texttt{github.com/HKU-MMLab/MIST}}\\
  \end{tabular}
  \vspace{0.55em}
\end{center}

\section{Introduction}
\label{sec:introduction}

\begin{figure}[H]
\centering
\includegraphics[width=\linewidth]{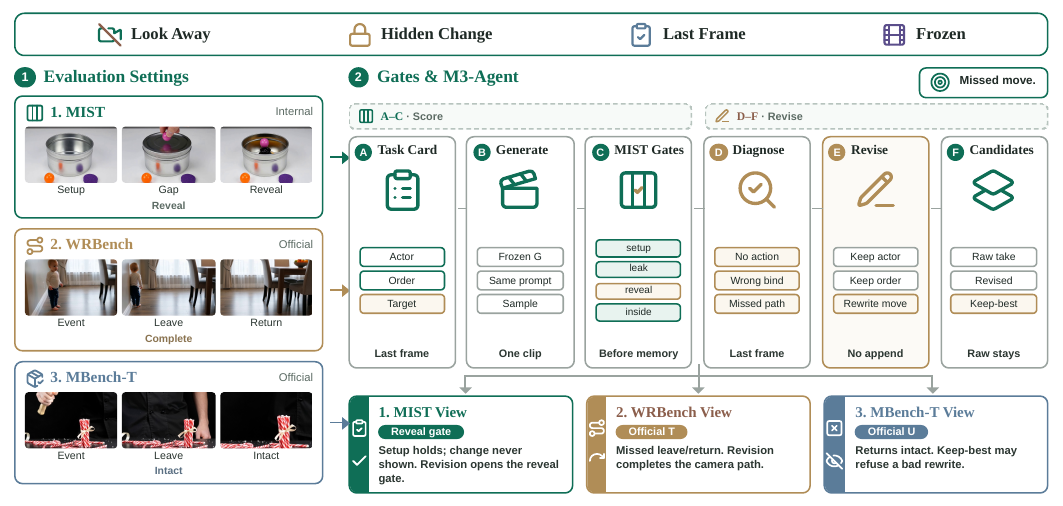}
\caption{\textbf{Execution before memory.}
Left: the three settings, each a clip in which the camera looks away
before the change is due.
Right: MIST scores the gates of the reveal; M3-Agent reads the last
frame, names the failed move, and revises only that move on the same
frozen generator.
Bottom: what each setting shows once the gates are scored.}
\label{fig:overview}
\end{figure}

Memory tests for video world models ask about a scene the camera no
longer shows: who sat down, what went into the box, whether the
candy broke
\citep{brooks2024videoworld,bruce2024genie,valevski2024gamengen}.
Object relations have to survive when the camera looks away, and
actions have to update them
\citep{shin2026unobserved,ma2026stevo,chen2026memobench}.
Current tests fold these checks into one number: a person and a
chair, an object left in a room, or a cube still in a cup
\citep{lu2026wrbench,zhang2026mbench}. A low number is easy to read
as forgetting, yet it cannot say whether the scene was set, whether
the asked change was carried out, or whether the last frame shows
the result (Fig.~\ref{fig:overview}).

\begin{figure}[t]
\centering
\includegraphics[width=\linewidth]{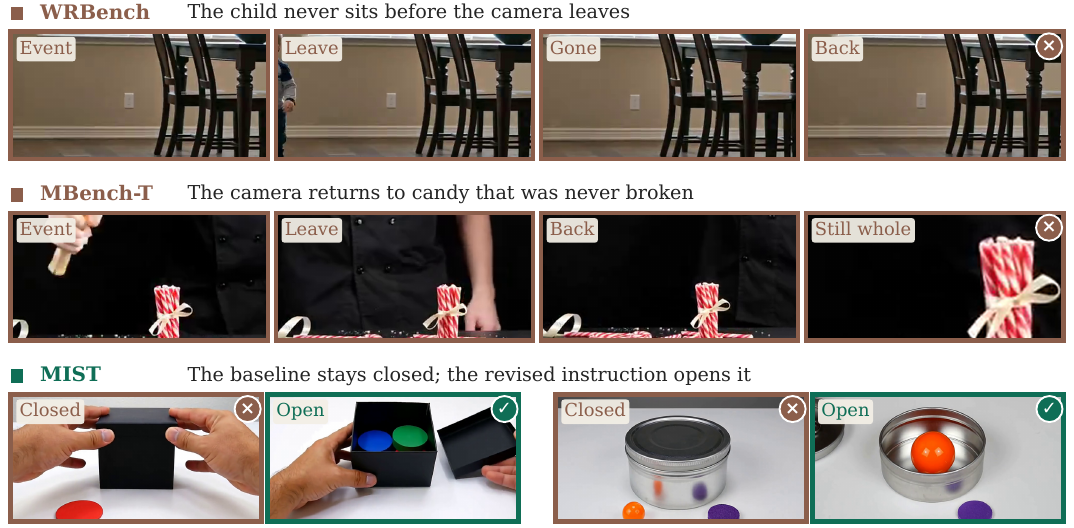}
\caption{\textbf{The last frame does not show the asked-for result.}
Three settings, not one run.
(Top)~WRBench: the child walks off before sitting, so the chair is
empty when the camera comes back.
(Middle)~MBench-T: the camera returns to candy that is still whole.
(Bottom)~MIST: two baseline clips that stay closed, each next to the
same case after the instruction is revised, where the box opens onto
the hidden objects.}
\label{fig:cases}
\end{figure}

When the change is not carried out, the hidden transition cannot
appear, so the test is not really a test of memory.
Figure~\ref{fig:cases} shows the usual clip. The child never sits
before the camera leaves. The camera comes back to candy that was
never broken. A box stays shut. The task names the target, but the
last frame never shows it. The score drops before memory is tested.
Until that change happens, asking the model to remember targets the
wrong stage. Tools that rewrite a video for quality, physics, or
preference do not ask whether the asked change was shown
\citep{lee2024videorepair,xue2024phyt2v,liu2025videot1}.
What is needed first is an evaluation that separates the change
from its result.

We introduce MIST to separate the two. The model first
sets the scene while the change stays hidden, and only later is
asked to show it, with leakage before that point kept under
control. On Wan TI2V-5B, three 5-second shots and 43 cases, the
scene is usually set and the hidden change does not leak, yet 41 of
43 baseline clips never show it. Asking only to open leaves most
lids down, and naming the answer helps less than revising the
failed move (\S\ref{sec:preexp}).
Looking at the last frame and revising the instruction makes the
change appear, though the shown state can still be wrong. The
bottleneck is therefore execution: the scene is set, but the
instructed change is not carried out.

The same gap appears in general memory evaluations. Failed clips on
WRBench skip the departure or the return, and failed clips on
MBench-T return to an object that is still intact
(\S\ref{sec:results}). On that basis we propose M3-Agent
(Make the Missing Move). It recursively revises the
details of the input instruction so that the same model carries out
the requested instruction. No memory module is added or trained. Relative to the baseline,
MIST reveal rises from 0.147 to 0.492. On the benchmarks,
WRBench's score $T$ rises by 26.1\% (0.452$\to$0.570, $n{=}40$) and
MBench-T's score $U$ by 52.7\% (0.292$\to$0.446, $n{=}30$; 23 of 30
keep the baseline).

These results say where the score is lost. They do not show that
the model stored the target state. A judged last frame is not
stored memory (\S\ref{sec:discussion}). They do show that the model
can already carry out many missing changes once the instruction
names them. The method meets the problem from the execution side:
it shows what the frozen model can already do once the move is
stated.

This paper makes three contributions (Fig.~\ref{fig:overview}).
\textbf{(1)}~MIST, an evaluation that scores whether the scene is
set, whether the hidden change leaks, and whether it is finally
shown, so a missing change is not reported as forgetting
(\S\ref{sec:benchmark}--\ref{sec:preexp}).
\textbf{(2)}~Evidence that failed official clips also miss the asked
change: a skipped return on WRBench, an intact object on MBench-T
(\S\ref{sec:results}).
\textbf{(3)}~M3-Agent, which recursively revises the input
instruction on the same model. It more than triples MIST reveal and
raises $T$ by 26.1\% and $U$ by 52.7\% (\S\ref{sec:m3},
\S\ref{sec:results}).

\section{Related Work}
\label{sec:related}

A low memory score can come from failed persistence, a wrong
state update, or a missing operation that never makes the state
observable. The three lines of work below each report one total.

\thinparagraph{Video generators as world models.}
Large video generators are treated as simulators or interactive
environments~\citep{brooks2024videoworld,bruce2024genie,
alonso2024diamond,valevski2024gamengen,huang2025vid2world}.
Long-horizon systems add packing, retrieval, or persistent state to
reduce drift~\citep{henschel2024streamingt2v,zhang2025framepack,
yu2025contextmemory,hong2025relic,wu2026infiniteworld,
chen2026reworld}. Neither line shows that a hidden state was kept
and updated.

\thinparagraph{Hidden state and stage-resolved evaluation.}
A hidden state cannot be scored if the requested update never
happens. StEvo-Bench, MemoBench, and the Video Shell Game study
updates under occlusion or unseen
swaps~\citep{shin2026unobserved,ma2026stevo,chen2026memobench}.
WRBench and MBench add a camera leave and a later
check~\citep{lu2026wrbench,zhang2026mbench}. Temporal and physical
suites likewise separate whether a transition occurs from whether
its endpoint is correct~\citep{feng2024tcbench,bansal2024videophy,
li2025worldmodelbench}. MIST applies that principle to delayed
reveal; it does not replace Official WRBench, Official MBench, or
the Video Shell Game.

\thinparagraph{Judges and test-time refinement.}
Structured VLM judges help on appearance, but remain weakest on
temporal and motion errors~\citep{tong2024cambrian,yang2025videogeneval}.
VideoRepair, PhyT2V, Video-T1, and VQQA already diagnose or rewrite
videos for quality, physics, or preference
\citep{lee2024videorepair,xue2024phyt2v,liu2025videot1,song2026vqqa},
but none asks whether the asked change was shown. M3-Agent instead
revises the instruction so the model carries that change out. We
claim no numbered comparison with the tools above. The broader map
is in \cref{app:literature-map}.

\section{MIST: separating scene setup from delayed state reveal}
\label{sec:benchmark}

\begin{figure}[t]
\centering
\includegraphics[width=\linewidth]{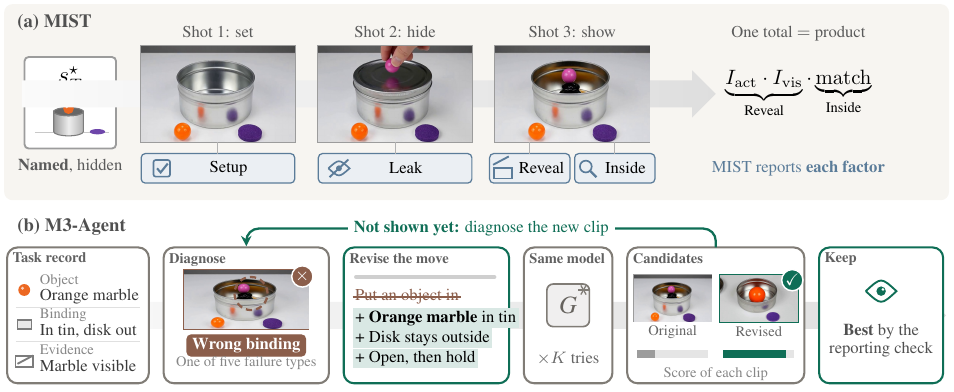}
\vspace{2pt}
\caption{\textbf{MIST scores each gate of the reveal; M3-Agent
revises the failed move until the change is shown.}
One Internal illustration. Generators in this paper include Wan
TI2V-5B, Wan T2V-A14B, and MiniMax-H3
(\cref{sec:experiments,sec:other-cells}).
(a)~Setup and leak check the first two shots; reveal and inside split
the last shot into the factors of Eq.~\eqref{eq:score}. The raw take
opens the tin onto another object.
(b)~Diagnose the last frame, revise the move rather than the answer,
and generate again with the same model. The rewrite puts in the orange
marble and opens to it; the original stays a candidate.}
\label{fig:method}
\label{fig:gates}
\label{fig:wording}
\label{fig:h3-strip}
\end{figure}

The cases in Fig.~\ref{fig:cases} show a missing final state, but
not why it is missing. A closed lid and a change that was never
carried out can produce the same total. MIST separates the two
(Fig.~\ref{fig:method}a). The model first sets the scene while
the change stays hidden, and only later is asked to show it.

\thinparagraph{What the evaluation does.}
A delayed query asks who sits where, what is inside, or how many
objects remain. In text, this is a question about a target state
$s_T^\star$. For a video generator, the clip must also produce that
state on screen. MIST names $s_T^\star$, hides it behind an opaque
gap, and only then asks the generator to show it. Each case is three
5-second shots: the first establishes the scene, the second passes
through the opaque gap, and the third requests the reveal. All
reveal comparisons reuse the same locked first two shots. The
baseline prompt and the gap never contain the answer; the ablations
that do name it are marked as such.
\label{app:information-flow}
The paired reveal analysis freezes the first two shots and regenerates
only the last shot from the same pre-reveal context.
Answer-only and diagnosis-guided arms may place the target in the
reveal prompt; Table~\ref{tab:repair} marks that access.
A prompt-channel audit confirms the target in the M3 and same-sentence
templates and confirms its removal from the generic open-and-hold arm,
so that arm is answer-free.
The same-sentence resample is an equal-compute control, not a
target-free control.
R0--R2 cover presence and count, binding, and history-update
operations (Table~\ref{tab:mist-families}).
\label{app:mist-families}

\begin{table}[H]
\caption{\textbf{MIST varies both the target state and the operation
needed to reveal it.}}
\label{tab:mist-families}
\centering
\small
\begin{tabular}{@{}lll@{}}
\toprule
 & \textbf{Target state} & \textbf{Required visible operation} \\
\midrule
R0 & Presence, absence, count, final state &
Open, reveal, or hold the right objects \\
R1 & Object--container or left--right binding &
Place the object in the named slot or box \\
R2 & History update: swap, remove, transform &
Execute and reveal the \emph{updated} state \\
\bottomrule
\end{tabular}
\end{table}

\thinparagraph{Three observable gates.}
Let $x_{1:T}=G(\pi,c)$ denote a clip from a frozen generator
$G$~\citep{wan2025}, prompt $\pi$, and image or caption context $c$,
and let $J$ be a fixed VLM judge~\citep{liu2023llava,tong2024cambrian}.
We distinguish whether the required move occurs
($I_{\mathrm{act}}$), whether the last frame is readable
($I_{\mathrm{vis}}$), and whether that frame matches the target
($\mathrm{match}(x,s_T^\star)=J(x_T,s_T^\star)$). A score that
requires all three is
\begin{equation}
\mathrm{score}(x,s_T^\star)
\;=\;
I_{\mathrm{act}}(x)
\cdot
I_{\mathrm{vis}}(x)
\cdot
\mathrm{match}(x,s_T^\star).
\label{eq:score}
\end{equation}
Eq.~\eqref{eq:score} says that a total collapses when any factor is
missing, and the total alone cannot tell a missing action from an
incorrect state. We therefore report the factors separately rather
than their product (Fig.~\ref{fig:method}a): setup checks the first
shot, leak the gap, and reveal and inside the last shot.
Never-open is the thresholded action class, not the mean reveal: a
low mean is not a never-open count (\cref{app:metric-accounting}).
A valid setup followed by a closed lid places the failure on the
reveal path, not on forgetting.

WRBench and MBench-T test the same gates in one continuous clip.
Each publishes a \emph{unique} score, $T$ and $U$, that gives no
credit when a step of this path is missing:
\begin{equation}
a \;\to\; \mathrm{leave}\;\text{until unjudgeable}
\;\to\; \mathrm{return} \;\to\; \mathrm{hold}(x_T),
\label{eq:unique}
\end{equation}

where $a$ is the requested event, such as a sit or a break. The
score thus requires the executed move, not only a final match
(\S\ref{sec:results}).

\section{Pre-experiments: the target is named, but the reveal is missing}
\label{sec:preexp}
\label{sec:diagnosis}
\label{sec:repair}

We apply the MIST measures before designing an agent or evaluating
an official suite. Two questions come first. At which gate does the
baseline fail, and can naming the target fix it?

\thinparagraph{Does the scene fail before memory is tested?}
Setup is high (Table~\ref{tab:diagnosis}): in most first shots the
objects and the container are in place. In 38 of 43 cases setup is
at least 0.8 while the baseline reveal stays below 0.3. What is missing is the later move that shows the asked-for
state.

\thinparagraph{Where does the baseline fail?}
Mean reveal stays low. When each clip receives one label, 41 of 43
baseline clips never open; one opens to an empty or blocked view,
and one is correct. In the closed cases the final state cannot
be evaluated at all, because the required action never exposes it.
On those 41 clips, a hand comes in and does not open in 38, nobody
comes in in 3, and the wrong object is opened in none. The shot
length is not why the lid stays down (\cref{app:residual-failures}),
but the three-shot stitch shapes how the failure looks: on a packed
15s MiniMax-H3 take the same judge finds 3 of 43 never-open and 24
of 43 opened onto the wrong state (\cref{app:transfer-mist43}). The
failure moves from not opening to opening wrong; in both cases the
asked change is not carried out.
A low total here says nothing about whether the model forgot
$s_T^\star$: persistence has not been tested yet.

\thinparagraph{Did the hidden gap leak the answer?}
No target is visible in the middle chunk.
\label{app:leak}
Paired gap leak is $0/43$ for both the baseline and the later rewrite.
A broader 51-row baseline export also contains no leak, but that export
is not the paired denominator. Setup is not a leak. The reveal failure
does not follow an early exposure of the answer; it occurs at the
action gate in Eq.~\eqref{eq:score}.

\begin{figure}[t]
\centering
\includegraphics[width=\linewidth]{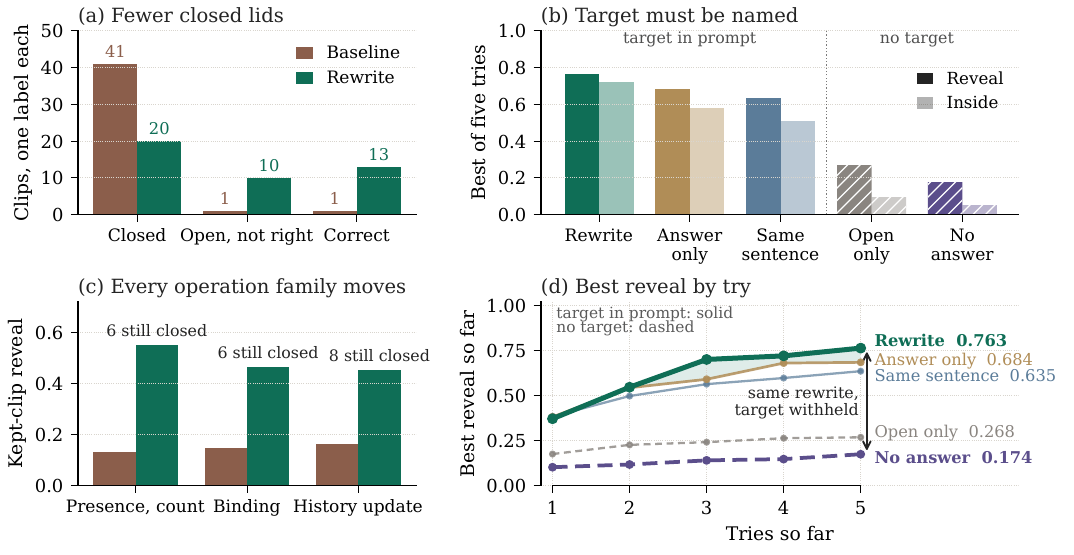}
\vspace{2pt}
\caption{\textbf{Rewriting the failed move opens more lids, but only
when the target is named.}
(a)~Failure classes for one kept clip per case
(Table~\ref{tab:diagnosis}).
(b)~Best-of-five reveal (dark) and inside (light) for the five arms
of Table~\ref{tab:repair}; hatched arms have no target in the prompt.
(c)~Kept-clip reveal by operation family, with the lids that stay
closed after the rewrite.
(d)~Best reveal so far as the five-try budget grows; dashed lines have
no target; the arrow marks the same rewrite without the target.}
\label{fig:story}
\label{fig:pass-budget}
\label{fig:selection}
\end{figure}

\begin{table}[t]
\caption{\textbf{The rewrite changes the reveal gate, not the setup.}
One kept clip per case, not best-of-five. Decimals are mean judge
scores; counts use one ordered label per clip (closed $\succ$ empty
$\succ$ wrong $\succ$ correct). Inside match need not reveal; Correct
must.}
\label{tab:diagnosis}
\centering
\small
\begin{tabular}{@{}lccc@{}}
\toprule
 & \textbf{Base} & \textbf{Rewrite} & $\boldsymbol{\Delta}$ \\
\midrule
Setup & 0.852 & 0.851 & $-0.001$ \\
\rowcolor{bestrow}
Reveal & 0.147 & \textbf{0.492} & \textbf{+0.344} \\
Inside match & 0.047 & \textbf{0.442} & +0.395 \\
\midrule
Never opens & 41 & 20 & $-21$ \\
Empty / blocked & 1 & 0 & $-1$ \\
Open, wrong & 0 & 10 & +10 \\
Correct & 1 & \textbf{13} & +12 \\
\bottomrule
\end{tabular}
\end{table}
\thinparagraph{What changes after a targeted rewrite?}
We inspect the last shot, diagnose its failed move, and rewrite only
that shot. With one kept clip per case, rather than best-of-five,
reveal rises and inside match rises with it
(Table~\ref{tab:diagnosis}). Setup remains flat.
Per case, reveal rises in 35 of 43, stays the same in 6, and falls
in 2; the mean paired change is 0.34 (95\% CI 0.24 to 0.45).
The failure class also changes. Of the baseline never-opens, rewrite
leaves many closed, opens some onto a wrong state, and makes some
correct (Fig.~\ref{fig:story}a). The table's open-but-wrong clips
include newly opened wrong states plus the initially correct case
\texttt{R2-10}, which the rewrite can break. The rewrite exposes more
states, but it does not guarantee that the exposed state is correct.
Accounting detail is in \cref{app:metric-accounting}.

All three operation families move the same way, and setup stays
flat in each: reveal goes from 0.132 to 0.549 on presence and count,
from 0.147 to 0.464 on binding, and from 0.163 to 0.454 on a history
update (Fig.~\ref{fig:story}c; \cref{app:family-stage}). Closed lids
that remain are not piled in one family. Openings onto the wrong state
are failed updates, not noise in one family
(\cref{app:residual-failures}). The rewrite has to spell out both the
move and the last-frame evidence, not only a verb.

\finding{1}{A hidden state cannot be judged until the model makes it
visible; a visible opening can still expose the wrong state.}

\thinparagraph{Is naming the target enough?}
The kept-clip diagnosis above locates the failure at the reveal gate;
it does not yet explain what kind of instruction opens that gate. If
the missing piece were only the answer, inserting $s_T^\star$ should
suffice. If it were only a verb, ``open and hold'' should suffice. We allocate the
same five tries in five ways (Table~\ref{tab:repair},
Fig.~\ref{fig:story}): the answer sentence is present or absent, and
the failed move is either rewritten or left unchanged. A fifth arm
rewrites the move from the same last frame but forbids every target
noun, color, and count (audited; 0/43 leaks).

The ordering in Table~\ref{tab:repair} separates these explanations.
The bare ``open and hold'' instruction performs worst. Supplying the
answer helps, but resampling the same sentence does not match a
diagnosis-guided rewrite. Looking at the clip and changing
the failed move gives the strongest best-of-five reveal and inside
scores (0.763 against 0.684 for answer-only). With the target
withheld the same rewrite reaches 0.174 and leaves 28 of 43 lids
closed: on this probe the rewrite opens the lid only when the prompt
also says what is inside. Each arm keeps, per case, the try whose
last frame the judge scores highest, so no cell here is the one-clip
mean of Table~\ref{tab:diagnosis}. Later tries are not better on
average, so the best-of-five cell is a running best
(Fig.~\ref{fig:story}d).
Where the same cases have an unchanged-prompt control, extra draws of
the same sentence do not explain the gain. The running best moves most
between the second and third try, from 0.546 to 0.700, and the last
two tries add little.
On six cases with a matched resample, the same-prompt running best is
$0.410$, $0.557$, $0.561$, $0.570$, $0.580$, while the rewritten move
on those same cases reaches $0.884$.
Constraints-only and ledger-only reach $0.189$ and $0.215$ on that
subset. The subset is too small to replace the 43-case table.
The pass-by-pass record behind Fig.~\ref{fig:story}d is in
\cref{tab:mist-pass-complete}: per-pass means are not monotone, and
``inside of chosen'' is not the kept-clip inside match $0.442$.

\begin{table}[b]
\caption{\textbf{With five tries, the rewrite helps only when the
answer is also in the prompt.}
Same judge; each arm keeps the try with the highest last-frame score.
Closed / Ok count never-open and fully correct kept clips.
Look-then-rewrite beats answer-only; the same rewrite with the
target withheld falls below open-and-hold.}
\label{tab:repair}
\label{tab:main-triptych}
\centering
\small
\setlength{\tabcolsep}{5pt}
\begin{tabular}{lccccc}
\toprule
\textbf{Arm} & \textbf{Ans.?} & \textbf{Reveal} & \textbf{Inside} &
\textbf{Closed} & \textbf{Ok} \\
\midrule
\rowcolor{bestrow}
Look, then rewrite & yes & \textbf{0.763} & \textbf{0.721} & 5 & \textbf{29} \\
Answer only & yes & 0.684 & 0.581 & 3 & 24 \\
Same sentence & yes & 0.635 & 0.512 & 4 & 21 \\
Open and hold & no & 0.268 & 0.093 & 6 & 3 \\
Rewrite, no answer & no & 0.174 & 0.047 & 28 & 2 \\
\bottomrule
\end{tabular}
\end{table}

\finding{2}{Rewriting the failed move beats naming the target alone,
but only when the target is also named: with it withheld, the same
rewrite opens almost nothing.}

The same missing operation appears on an Internal MiniMax-H3 take
(Fig.~\ref{fig:wording}): the raw take puts a different object in the
tin and opens to it; the rewritten instruction puts the orange marble
and opens to that. Clip settings and cell boundaries are in
\cref{sec:other-cells}.

Asking only to open fails, naming the target helps, and revising the
failed move helps most, but only with the target named. The next
section turns look, diagnose, and change into a test-time workflow.

\section{M3-Agent: revise the instruction for the missing change}
\label{sec:m3}

The pre-experiments show that the workflow must revise the failed
instruction, not append the target or repeat the same prompt.
M3-Agent (Make the Missing Move) does that at test
time (Fig.~\ref{fig:method}b): it reads the last frame, revises the
details of the instruction for that change, and generates again with
the same model. It does not add or train a memory module.

The workflow has three steps. First, a short task record lists the
actor or object, its required binding, the operation, and the
evidence that must appear in $x_T$; these fields come from the task
card. Second, a VLM inspects the generated last frame against that
record and assigns a visible failure mode, such as no action, wrong
binding, missing departure, return to an intact object, or an
occluded view. Third, the workflow rewrites the failed operation
rather than the answer, and the frozen generator produces another
candidate. This third step is the one that works in
Table~\ref{tab:repair}.
The rewrite preserves the actor, object, target state, and temporal
order; only the failed operation and its final-frame evidence clause
may change. Diagnosis reads the rendered \(x_T\) once, never the
generator's hidden states.

\thinparagraph{A worked case.}
Figure~\ref{fig:method}b traces the workflow on that case. The
task record asks for the orange marble in the tin, the purple disk
outside, and a visible marble at the end. The raw take opens the tin
onto another object, which the record labels a wrong binding rather
than a missing action. The revised instruction keeps the tin, the
disk, and the order of events, and names the object that must go in;
the next take opens onto the orange marble. The raw take stays in the
candidate set, so a revision that makes things worse is not kept.

Given $K$ candidates, M3-Agent keeps the clip that best carries out
the instruction,
\begin{equation}
\hat{x}
\;=\;
\underset{k\le K}{\arg\max}\;
\mathrm{score}\!\left(x^{(k)},s_T^\star\right),
\label{eq:keep}
\end{equation}

and the original candidate remains in the set. The score here is the
same check that reports the result: the MIST judge on MIST, and the
official gates on WRBench and MBench-T. A reader that sees only the
last frame, without that score, does not reliably beat the baseline
(\S\ref{sec:discussion}). A single revision is not always better:
\texttt{R2-10}, for example, turns a correct case into an incorrect
one, so the original clip stays eligible. Searching several
instructions at test time is a known strategy
\citep{snell2024testtime,oshima2025dlbs,he2025evosearch,
khalaf2025rewardhacking,landesberg2026judges}.
Every reported best-of-$K$ and official keep-best cell is chosen after
generation with the reporting scorer; that is an oracle coverage
estimate, not a deployable critic
(\cref{app:m3-implementation}).

\thinparagraph{On MIST.}
The agent works with frozen TI2V-5B. It preserves the first two
5-second shots and rewrites only the final reveal shot, and the
reveal prompt may include $s_T^\star$ (Table~\ref{tab:repair}). The
kept-clip analysis in Table~\ref{tab:diagnosis} locks one outcome per
case; Table~\ref{tab:repair} instead generates $K{=}5$ candidates per
arm and applies Eq.~\eqref{eq:keep}.

\thinparagraph{On WRBench and MBench-T.}
The agent works on 10-second clips from frozen T2V-A14B and keeps
each suite's official camera clause; on WRBench the caption stays
\texttt{task-context none}. Candidate instructions establish the
requested relation before departure and hold it after return. On
MBench-T the candidate set must include the original clip: forcing
another leave--return can preserve the camera motion while worsening
the returned state. Selection is made per sample over a small
instruction set.

\section{Evaluation setup}
\label{sec:experiments}

We never merge numbers from different scorers. Every number below is
\textbf{Official} (the suite's own scorer) or \textbf{Internal} (our
fixed judge). Adapted protocols are not pooled with these rows. The
three-chunk MBench condition is omitted.
Main claims use Wan TI2V-5B for Internal MIST, Wan T2V-A14B for
Official WRBench and MBench-T, and MiniMax-H3 as an Internal
qualitative pilot (Fig.~\ref{fig:wording}; \cref{sec:other-cells}).
Clip lengths, seeds, judges, and budgets are in
\cref{tab:setup,tab:protocol-cards,tab:mist-pass-complete,tab:app-selector}.
Internal and Official exports keep separate manifests and are never
pooled.

\section{Main results: the same diagnosis on official scores}
\label{sec:results}

Under the scorers fixed in \S\ref{sec:experiments}, we ask whether
official memory protocols fail at the same execution gates isolated
by MIST. Inspecting failed clips on official WRBench and MBench-T
shows the same missing action. The analysis transfers a diagnosis,
not a number: Internal MIST, Official unique scores, and packed H3
pilots remain separate cells (\cref{sec:other-cells}).

\thinparagraph{What do the official scores contain?}
Both suites already require the path in
Eq.~\eqref{eq:unique}~\citep{lu2026wrbench,zhang2026mbench}.
We report which gates change and which clip is kept per item.
We do not treat a correlation between a total and one of its own
parts as a separate causal finding
(\cref{app:official-complete}).

\thinparagraph{What happens when we search that path?}
The two official protocols fail at different points, and those
points map onto the two remaining MIST failure types. On WRBench, the
camera often does not complete the required departure and return:
a shut $D6$ gate is the same class as a never-open, so
persistence is not yet measured. On MBench-T the camera returns,
but the object is often still intact (Fig.~\ref{fig:cases}): the
path completed, yet the aftermath is the wrong world state, as in
the open-but-wrong MIST clips. The search therefore has two different
jobs. On WRBench it has to find a missing path. On MBench-T it has to
keep the original clip when a fixed rewrite harms the aftermath.

\begin{table}[t]
\caption{\textbf{Official MBench-T: selection can refuse a harmful
rewrite.} Keep-best raises $U$ by 52.7\%; one clip per item gives
$0.443$. Trig.\ is the state/correctness fire rate. State here is
not MIST reveal.}
\label{tab:official-mbench}
\centering
\small
\begin{tabular}{lcccc}
\toprule
\textbf{Clip} & $\boldsymbol{U}$ & \textbf{State} & \textbf{Corr.} & \textbf{Trig.} \\
\midrule
Baseline & 0.292 & 0.350 & 0.233 & 0.93 / 0.83 \\
Force leave--return & 0.251 & 0.242 & 0.260 & 0.93 / 0.87 \\
Force last-frame lock & 0.228 & 0.208 & 0.247 & 0.97 / 0.97 \\
\rowcolor{bestrow}
Keep-best & \textbf{0.446} & 0.492 & 0.400 & 1.00 / 1.00 \\
\bottomrule
\end{tabular}
\end{table}

\begin{table}[t]
\caption{\textbf{Official WRBench: the search opens more return gates
than visual quality.} Kept clip raises $T$ by 26.1\%. $T$ zeros a
missing return. Best-looking differs from the $T$-kept clip on 24 of
40 items.}
\label{tab:camera-ablate}
\label{app:external}
\centering
\small
\begin{tabular}{lcccc}
\toprule
\textbf{Clip} & \textbf{Vis.} & \textbf{D5} & \textbf{D6} & $\boldsymbol{T}$ \\
\midrule
Baseline & 0.648 & 16/40 & 18/40 & 0.452 \\
One leave--return & 0.690 & 23/40 & 14/40 & 0.496 \\
Best-looking & \textbf{0.755} & 20/40 & 13/40 & 0.514 \\
\midrule
Search, first set & 0.681 & 29/40 & 23/40 & 0.550 \\
\quad + second set & 0.662 & 28/40 & 27/40 & 0.547 \\
\rowcolor{bestrow}
\quad + last step (kept) & 0.654 & \textbf{31/40} & \textbf{31/40} & \textbf{0.570} \\
\bottomrule
\end{tabular}
\end{table}

\begin{figure}[t]
\centering
\includegraphics[width=\linewidth]{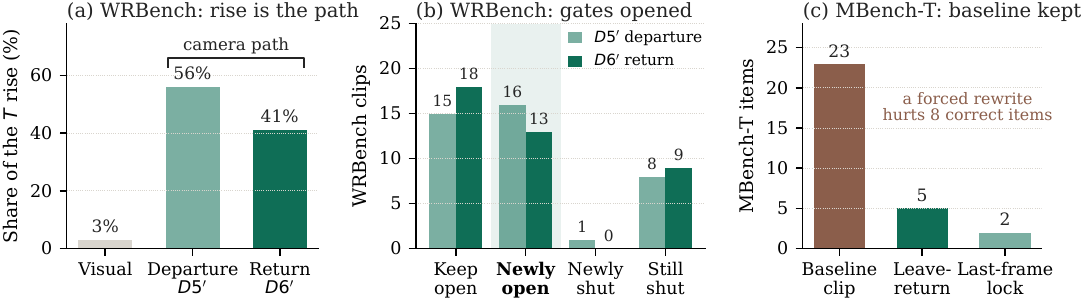}
\caption{\textbf{The same diagnosis has different consequences on
the two official suites.}
(a)~Share of the WRBench $T$ rise from visual quality, the departure
check $D5'$, and the return check $D6'$.
(b)~WRBench departure and return gates that stay open, newly open,
newly shut, or stay shut between the baseline and the kept clip.
(c)~MBench-T items whose kept clip is the baseline, the leave--return
instruction, or the last-frame lock. Panels keep separate
denominators.}
\label{fig:stats}
\end{figure}

On WRBench, score \(T\) rises (Table~\ref{tab:camera-ablate}) because
the camera finishes the return, not because the frame looks better.
The paired gain has a 95\% bootstrap interval of $+0.070$ to $+0.166$
(\cref{app:m3-implementation}).
The component audit attributes about 3--4\% of \(\Delta T\) to visual
quality, 55--56\% to the departure check \(D5'\), and 41\% to the
return check \(D6'\), depending on rounding (Fig.~\ref{fig:stats}a).
Of the 40 return gates, 18 stay open, 13 newly open, none close, and
9 stay shut (Fig.~\ref{fig:stats}b).
A second instruction set can lower the total, from \(0.550\) to
\(0.547\); the last step restores it to \(0.570\).
Adding a skill is therefore not monotone.
\cref{tab:app-selector} places the kept result next to two ceilings
that are not the method: a peek at the gate followed by visual
selection reaches \(0.586\), and the maximum \(T\) in the pack is
\(0.601\). The reported method is the unique-head result
\(T{=}0.570\).

On MBench-T the camera already comes back, so \(U\) follows the
state after return, not whether the camera left
(Table~\ref{tab:official-mbench}).
Fourteen of 30 cases stay hard: the leave--return trigger usually
fires, but the object comes back unchanged or wrong, and a prompt
that only says to hold the update fixes none of those 14.
Fixed leave--return hurts 8 cases whose baseline state score is 1.0.
A last-frame lock, used as the only clip, gives \(U=0.228\); it helps
the pack only because selection may refuse it.
A later destroy-and-hold instruction stays near the floor as a fixed
arm and flips one additional case only under oracle selection.
That cell is not a headline method.
Selection keeps the baseline on 23 items, the leave--return
instruction on 5, and the last-frame lock on 2
(Fig.~\ref{fig:stats}c).
The published \(0.446\) averages a state pick and a correctness pick.
One clip per item gives \(0.443\).
Trigger versus conditional scores are in
\cref{tab:mbench-official-details}; they are not MIST metrics.

\thinparagraph{What do the ablations constrain?}
\label{sec:ablations}
Table~\ref{tab:repair} rules out answer text or unchanged
resampling as the sole explanation for the MIST rewrite. The
official ablations add two boundaries. On MBench-T, a prompt that
only asks to hold the update does not fix the hard cases, and a
last-frame lock alone is worse than the baseline
(Table~\ref{tab:official-mbench}). On WRBench, the best-looking clip
and the clip kept for $T$ are different clips
(Table~\ref{tab:camera-ablate}), so a readable frame is not the
executed path. Score-composition plots are in
\cref{fig:suite}.

\begin{table}[t]
\caption{\textbf{A judge can name the target from text before the
video shows it.} YES rates. The top block is text only; the bottom
block is the locked last frame. The 122B text-only cells were not run.}
\label{tab:who}
\centering
\small
\begin{tabular}{@{}lrr@{}}
\toprule
\textbf{Input} & \textbf{122B} & \textbf{8B} \\
\midrule
Gold text only & n/a & 0.91 \\
Setup caption only & n/a & 0.40 \\
Shuffled gold only & n/a & 0.07 \\
\midrule
Last frame, baseline & 0.023 & 0.09 \\
\rowcolor{bestrow}
Last frame, rewrite & \textbf{0.442} & \textbf{0.56} \\
\bottomrule
\end{tabular}
\end{table}

\thinparagraph{Can the judge answer without the video?}
\label{sec:who}
A text-only control separates a text shortcut from visual evidence
(Table~\ref{tab:who}). A held-out 8B reader names gold $s_T^\star$
from text, then collapses when the gold sentence is removed or
shuffled. On the locked last frame it tracks the 122B inside check:
almost no baseline clip is accepted, and the rewrite rate matches
Table~\ref{tab:diagnosis}. Naming the target is easy; seeing the
executed state is not. Agreement with official gates is much lower
(\cref{app:judge}); the reread does not replace either official
scorer
\citep{han2025videobench,liu2025uve,chen2024mjbench,jia2026raveneval}.

\finding{3}{On both official suites the score is lost before
persistence is tested: WRBench clips skip the departure or return,
and MBench-T clips return to an object that is still intact.}

\finding{4}{Text can identify the target without establishing that
the video shows it; the last-frame check improves only after the
failed move is rewritten.}

\section{Discussion: what execution evidence can establish}
\label{sec:limitations}
\label{sec:discussion}

MIST separates the scene, the change, and the last frame
(Eqs.~\eqref{eq:score}--\eqref{eq:unique}); M3-Agent revises the
instruction. Neither shows what the model stored.

\thinparagraph{Measurement.}
A high reveal is not a memory claim. If the asked change never
occurs, persistence is untested, whether the clip stays closed or
opens onto the wrong state (Table~\ref{tab:diagnosis};
\S\ref{sec:preexp}).

\thinparagraph{Selection.}
Keep-best is an oracle over a small candidate pool, scored by the
same check that reports the result. The rewrite may name the
target, and a target-withheld control fails
(Table~\ref{tab:repair}). A last-frame reader that does not see the
score does not beat the baseline
(\cref{tab:app-blind-selector}). Because a rewrite can damage a
correct case, the original clip stays eligible.

\thinparagraph{Scope.}
The main study uses one generator family under Internal and Official
protocols; packed MiniMax-H3 cells stay Internal
(Fig.~\ref{fig:wording}; \cref{sec:other-cells}). The held-out judge
check is not human validation (\cref{app:judge}). Claim boundaries
are in \cref{tab:claim-boundary}.

\thinparagraph{What a positive result would need.}
We cannot claim the backbone stored \(s_T^\star\). That claim would
need a target-blind selector that beats the baseline, held-out
cases, and human labels on the exposed state. Until then, an opened
lid shows that the state can be judged, not that it was remembered.

\section{Conclusion}
\label{sec:conclusion}

MIST splits one memory score into three questions: was the scene
set, was the asked change carried out, and does the last frame show
the result. On the Internal cases the scene and gap usually hold,
yet the baseline often never shows the change, so persistence is
not yet tested. Revising the failed instruction exposes more of the
target state than naming it alone, though the shown state can still
be wrong.

M3-Agent revises the instruction so the same frozen model carries
out the change. Relative to the baseline, reveal rises on Internal
MIST and the Official unique scores improve, with opposite keep-best
consequences on the two suites. These cells do not establish that
the backbone stored \(s_T^\star\). A memory evaluation should
publish the three checks apart, not one total.
\label{sec:conclusion-end}

\bibliographystyle{plainnat}
\bibliography{mist,recent_verified}

\appendix
\section{Protocol cards and budgets}
\label{app:protocols}
\label{app:settings}
\label{app:reproducibility-details}
\label{app:budget}
\label{app:pass-budget}
\label{app:official-complete}
\label{app:m3-implementation}
\label{app:claim-boundary}
\label{app:step1}

Table~\ref{tab:setup} lists the settings behind the main claims.
H3 packed-15s pilots are a fourth Internal cell, not a fourth official
track (\cref{sec:other-cells}).
Table~\ref{tab:protocol-cards} expands each track.

\begin{table}[H]
\caption{\textbf{Three settings, three scorers.} $T$ averages WRBench
$D3$, $D4$, $D5'$, $D6'$ (leave/return); $U$ averages MBench-T state
and correctness. A missing return or untriggered clip scores zero.}
\label{tab:setup}
\centering
\small
\setlength{\tabcolsep}{4.5pt}
\begin{tabular}{@{}lllcll@{}}
\toprule
\textbf{Setting} & \textbf{Generator} & \textbf{Clip} & $\boldsymbol{n}$ & \textbf{Scorer} & \textbf{Reported} \\
\midrule
MIST & Wan TI2V-5B & $3{\times}5$s & 43 & Internal & setup / leak / reveal / inside \\
H3 pilot & MiniMax-H3 & packed 15s & 43 & Internal & failure classes (qual.) \\
WRBench & Wan T2V-A14B & 10s, \texttt{yaw\_LR} & 40 & Official & $T=(D3{+}D4{+}D5'{+}D6')/4$ \\
MBench-T & Wan T2V-A14B & 10s, 5 captions & 30 & Official & $U=(\mathrm{state}{+}\mathrm{corr})/2$ \\
\bottomrule
\end{tabular}
\end{table}

MIST uses Wan2.2-TI2V-5B, three 5-second shots, and seed class 9042.
The 122B judge reads six positions in the reveal shot,
\(\{0.90,0.93,0.95,0.97,0.99,0.995\}\).
Its prompt and parsing stay frozen across arms.
Paired intervals use 4{,}000 bootstrap resamples over the 43 cases
with seed 0; the paired Wilcoxon test uses the same pairs.
Official WRBench uses Wan2.2-T2V-A14B, 161 frames over 10 seconds,
\texttt{yaw\_LR}, \texttt{task-context none}, and the released scorer.
Official MBench-T uses the same generator and duration, five native
caption segments, and the released causal metrics.
These settings differ jointly from MIST, so TI2V-5B versus T2V-A14B
is not a controlled comparison of model scale.

\begin{table}[H]
\caption{\textbf{What each track actually runs.} Do not add a number
from one column into another.}
\label{tab:protocol-cards}
\centering
\small
\setlength{\tabcolsep}{4pt}
\begin{tabular}{@{}p{0.16\linewidth}p{0.26\linewidth}p{0.26\linewidth}p{0.26\linewidth}@{}}
\toprule
 & \textbf{MIST (Internal)} & \textbf{WRBench unique} & \textbf{MBench-T unique} \\
\midrule
Backbone & TI2V-5B, 3$\times$5s & T2V-A14B, 161f/10s & T2V-A14B, 161f/10s \\
Prompt & hide $s_T^\star$ in the gap & official \texttt{yaw\_LR} & native 5 captions \\
Gold in caption? & baseline no; rewrite arm yes & \texttt{task-context none} & no; a lock is an intervention \\
Judge & Qwen3.5-122B & official scorer & official \texttt{mbench} \\
Score & reveal / inside & $T=(D3{+}D4{+}D5'{+}D6')/4$ & $U=(\mathrm{state}{+}\mathrm{corr})/2$ \\
N/A & never-open is not forgetting & missing return scores 0 in $T$ & untriggered scores 0 in $U$ \\
Seed class & 9042 & 9042-class & 8042 (official split) \\
Execution head & last-shot rewrite, $K{=}5$ & pack search, two selectors & pack search; may refuse \\
\bottomrule
\end{tabular}
\end{table}

Table~\ref{tab:mist-pass-complete} is the pass-by-pass record behind
Fig.~\ref{fig:story}d. Per-pass means are not monotone. ``Inside of
chosen'' scores the clip selected for reveal; it is not the kept-clip
inside match $0.442$.

\begin{table}[H]
\caption{\textbf{Complete MIST running-best budget.}
Per-pass means are not monotone. ``Inside of chosen'' evaluates the
reveal-argmax clip and is not the kept-clip inside $0.442$.}
\label{tab:mist-pass-complete}
\centering
\small
\begin{tabular}{@{}cccccccc@{}}
\toprule
\(k\) & \textbf{Pass mean} & \textbf{Reveal best} &
\textbf{Inside of chosen} & \textbf{Closed} & \textbf{Wrong} &
\textbf{Correct} & \(\boldsymbol{\geq 0.6}\) \\
\midrule
1 & 0.371 & 0.371 & 0.256 & 19 & 14 & 10 & 10/43 \\
2 & 0.399 & 0.546 & 0.442 & 10 & 17 & 16 & 16/43 \\
3 & 0.484 & 0.700 & 0.628 & 8 & 10 & 25 & 25/43 \\
4 & 0.444 & 0.720 & 0.651 & 7 & 10 & 26 & 26/43 \\
5 & 0.560 & 0.763 & 0.721 & 5 & 9 & 29 & 28/43 \\
\bottomrule
\end{tabular}
\end{table}

Table~\ref{tab:app-selector} expands the WRBench search in
Table~\ref{tab:camera-ablate}. Peek-gate and max \(T\) are ceilings,
not the reported method.

\begin{table}[H]
\caption{\textbf{WRBench search budget.} Same clips as
Table~\ref{tab:camera-ablate}. Adding a second skill is not monotone.}
\label{tab:app-selector}
\label{tab:app-step1}
\centering
\small
\begin{tabular}{@{}lcccc@{}}
\toprule
\textbf{What we do} & \textbf{Vis.} & \textbf{D5} & \textbf{D6} & \(\boldsymbol{T}\) \\
\midrule
Camera only & 0.648 & 16/40 & 18/40 & 0.452 \\
Search over first skill & 0.681 & 29/40 & 23/40 & 0.550 \\
\quad ${+}$ second skill & 0.662 & 28/40 & 27/40 & 0.547 \\
\quad ${+}$ last step (kept) & 0.654 & 31/40 & 31/40 & 0.570 \\
One skill, no search & 0.690 & 23/40 & 14/40 & 0.496 \\
Peek gate, then vis. & 0.710 & 35/40 & 21/40 & 0.586 \\
Max $T$ (oracle ceiling) & 0.707 & 36/40 & 24/40 & 0.601 \\
\bottomrule
\end{tabular}
\end{table}

Table~\ref{tab:mbench-official-details} separates MBench-T trigger
rates from conditional scores. Those conditional scores are not MIST
metrics.

\begin{table}[H]
\caption{\textbf{Official MBench-T five-segment details.}
Trigger and conditional scores are separate. They are not MIST
metrics.}
\label{tab:mbench-official-details}
\centering
\small
\begin{tabular}{@{}lccc@{}}
\toprule
\textbf{Metric} & \textbf{Baseline} & \textbf{Fixed leave--return} &
\textbf{Keep-best} \\
\midrule
Prompt score & 31.528 & 31.382 & 32.303 \\
State trigger & 0.933 & 0.933 & 0.967 \\
State conditional & 0.375 & 0.259 & 0.474 \\
Correctness trigger & 0.833 & 0.867 & 0.967 \\
Correctness conditional & 0.280 & 0.300 & 0.352 \\
\bottomrule
\end{tabular}
\end{table}

Table~\ref{tab:app-blind-selector} re-selects one clip per item from
the same official packs without the official scorer. None of the
scorer-free selectors separates from the baseline.

\begin{table}[H]
\caption{\textbf{Official $T$ and $U$ when the selector cannot see
the scorer.} Same candidate clips as the main tables. Intervals are
paired differences from the baseline. Do not add across columns.}
\label{tab:app-blind-selector}
\centering
\small
\setlength{\tabcolsep}{4pt}
\begin{tabular}{@{}lcccc@{}}
\toprule
\textbf{Selector} & \textbf{WRBench $T$ ($n{=}40$)} & \textbf{95\% CI $\Delta$} &
\textbf{MBench-T $U$ ($n{=}30$)} & \textbf{95\% CI $\Delta$} \\
\midrule
First clip (baseline) & 0.452 & n/a & 0.292 & n/a \\
First rewrite only & 0.496 & $[-0.027, +0.114]$ & 0.251 & $[-0.172, +0.094]$ \\
Random, 20 draws & $0.487 \pm 0.019$ & $[-0.008, +0.076]$ & $0.265 \pm 0.030$ & $[-0.101, +0.027]$ \\
Blind 8B, last frame & 0.466 & $[-0.044, +0.070]$ & 0.248 & $[-0.107, +0.007]$ \\
Keep-best (sees scorer) & 0.570 & $[+0.070, +0.166]$ & 0.443 & $[+0.074, +0.243]$ \\
\bottomrule
\end{tabular}
\end{table}

\begin{figure}[H]
\centering
\includegraphics[width=\linewidth]{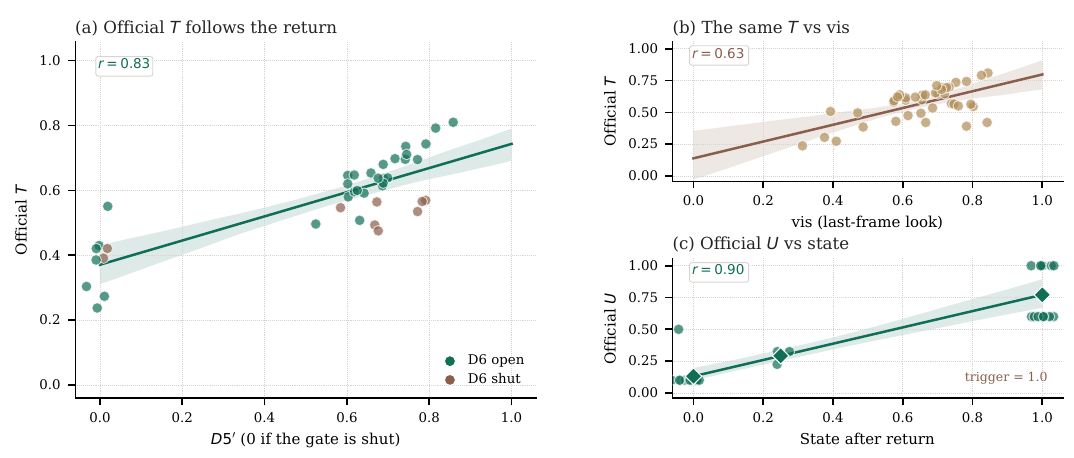}
\caption{\textbf{Official-score component plots.}
(a--b)~WRBench \(T\) against \(D5'\) and visual quality.
(c)~MBench \(U\) against returned state. Because these totals contain
the plotted components, the panels describe score composition. They
do not establish an independent causal correlation.}
\label{fig:suite}
\end{figure}

\label{app:metric-accounting}
\label{app:residual-failures}
\label{app:family-stage}
Of the 41 baseline never-opens, the rewrite leaves 19 closed, opens
nine to a wrong state, and makes 13 correct.
\texttt{R2-10} moves from a correct baseline to never-open, so the
exclusive never-open count in Table~\ref{tab:diagnosis} is 20 rather
than 19. Kept-clip reveal $0.492$ is not best-of-five $0.763$.
Baseline mean inside is $0.047$, while the strict 122B YES rate is
$1/43=0.023$; for the rewrite both round to $0.442$ ($19/43$).
Closed lids after rewrite: 6 / 6 / 8 on presence-count, binding, and
history update. The first pass revises 33 of 43 prompts.

Table~\ref{tab:claim-boundary} states the reading each main
observation supports, and the reading it does not.

\begin{table}[H]
\caption{\textbf{Evidence boundaries.} Each row is an observation
already reported in the main text. The middle column is the reading
used in this paper. The right column is not established.}
\label{tab:claim-boundary}
\centering
\small
\setlength{\tabcolsep}{3pt}
\begin{tabular}{p{0.28\linewidth}p{0.33\linewidth}p{0.33\linewidth}}
\toprule
\textbf{Observation} & \textbf{Supported reading} &
\textbf{Not established} \\
\midrule
Valid setup, clean gap, closed reveal &
The visible reveal path is missing &
The hidden state was forgotten \\
$41/43$ never-open, TI2V-5B $3{\times}5$s &
Reveal path missing in this Internal stitch &
Every generator or packed 15s take fails to open \\
Bo5 rewrite $0.763$ vs answer-only $0.684$; target withheld $0.174$ &
The rewrite helps only with the answer in the prompt; keep-best is an oracle &
Diagnosis separable from the target in the prompt \\
Blind 8B selector $T$ $0.466$, $U$ $0.248$ &
Scorer-free selection does not beat the baseline &
A deployable selector for either suite \\
Official keep-best $T$/$U$ rises &
Oracle coverage improves for that protocol &
A deployable critic, or one intervention that transfers \\
H3 $n{=}6$ Internal 8B reveal $1.000$ &
The same cards can be packed as 15s &
Official $T$/$U$, stored $s_T^\star$, or a scale claim \\
H3-43 continuity $0.605\!\rightarrow\!0.721$ &
Fewer hard cuts on packed 15s &
Reveal, inside, or Official $T$/$U$ \\
H3-43 122B reveal $0.387\!\rightarrow\!0.729$, never-open $3/43$ &
On a packed take the raw failure is a wrong state, not a closed lid &
That the stitch result transfers, or Official $T$/$U$ \\
\bottomrule
\end{tabular}
\end{table}

\section{Extended literature map and positioning}
\label{app:literature-map}

This section records the wider 2024--2026 literature used to bound
the paper's claims. Entries are grouped by the question they answer,
rather than presented as evidence for one aggregate notion of
``memory.''

\thinparagraph{What enables long and persistent rollouts?}
Diffusion and latent-video foundations establish the rendering
lineage~\citep{ho2020denoising,blattmann2023svd,wan2025}, while
physical-AI platforms broaden the world-model framing
\citep{yang2023cosmos}. Recent systems address exposure bias,
out-of-sight dynamics, bounded memory, and revisit stability through
self-conditioned training or explicit state
\citep{huang2025selfforcing,duan2026liveworld,alayaworld2026,
xu2026worldroam}. Omnimodal enterable worlds, controlled action-memory
probes, and long-horizon audio--visual story generation extend the same
line~\citep{zhang2026echowm,king2026echo,duan2026long}.
Agent-driven and embodied evaluations further test
whether action-conditioned worlds remain usable over long objectives
\citep{ding2026playworld,wang2026robograph,yang2026memorywam,
liu2026streampi}. EntityBench separately measures recurring entities
across multi-shot gaps~\citep{he2026entitybench}. Long-horizon
generators and packing systems also appear in the main related work
\citep{wu2024ivideogpt,xiang2024pandora,zhu2024irasim,
garcin2026persist,wang2026worlddirector}. These works motivate
long-horizon memory, but architectural success in one protocol does
not resolve MIST's execution-versus-persistence ambiguity.

\thinparagraph{Which benchmark dimensions are adjacent to MIST?}
WorldBench and IntPhys~2 isolate physical concepts and object
permanence, whereas EWMBench and VBench~2.0 separate scene, motion,
semantics, and intrinsic faithfulness
\citep{upadhyay2026worldbench,bordes2025intphys2,
hu2025ewmbench,zheng2025vbench2}. Understanding suites separate
endpoint answers from continuous
tracking~\citep{yu2026vstat,chen2026tocbench,han2026statetrace}.
LongVideoBench and MLVU test long-context understanding rather than
generator-side state
production~\citep{wu2024longvideobench,zhou2024mlvu}.
ChronoMagic-Bench separates amount of change from temporal coherence,
and T2VPhysBench evaluates physical laws with manual judgments
\citep{yuan2024chronomagic,guo2025t2vphysbench}. Adjacent temporal
and physical suites include TC-Bench, T2V-CompBench, OSC-Bench,
YoCausal, PhyGenBench, and VideoPhy2
\citep{sun2024t2vcompbench,han2026oscbench,xie2026yocausal,
meng2024phygenbench,bansal2025videophy2}. MIST instead fixes a
target state and asks whether the generated operation makes that
state judgeable after an opaque gap.

\thinparagraph{How has video generation been refined at test time?}
Reward gradients and human-feedback rewards improve video generation
through learned objectives
\citep{prabhudesai2024videoreward,liu2025videoreward}.
GenMAC and Prompt-A-Video use collaborative or preference-aligned
prompt revision~\citep{huang2024genmac,ji2024promptavideo}.
ScalingNoise, VISTA, RAPO++, and ImagerySearch broaden search over
noise, prompts, candidates, or evaluators
\citep{yang2025scalingnoise,long2025vista,gao2025rapopp,
wu2025imagerysearch}. LatSearch and noise-trajectory search move
selection inside diffusion inference, while GEARS diagnoses and
recycles candidates
\citep{zhao2026latsearch,ramesh2025noisesearch,he2026gears}.
M3-Agent is not a generic test-time search. Its contribution is an
operation-constrained response to a delayed-reveal diagnosis, and its
score-aware selection remains an oracle coverage analysis.

\thinparagraph{What can automated video evaluators establish?}
VBench and EvalCrafter established multidimensional evaluation beyond
single perceptual totals
\citep{huang2023vbench,liu2023evalcrafter}. VideoScore,
AIGV-Assessor, LOVE, and UVE learn or organize fine-grained
human-facing quality or task judgments
\citep{he2024videoscore,wang2024aigvassessor,wang2025love,
liu2025uve,li2024vlrewardbench}. VideoGen-Eval decomposes prompts
into checkable questions, while WorldJen and RAVEN-Eval study rubric
design, score compression, ordering, and confidence
\citep{yang2025videogeneval,inbasekar2026worldjen,jia2026raveneval}.
CLIP-style alignment and vision-centric VLMs provide useful readers
\citep{radford2021clip,liu2023llava,tong2024cambrian}, but no one
reader is treated here as ground truth for latent state. This is why
MIST reports gate-specific judgments, a held-out reread, and explicit
disagreement with official scorers.
Embedding-centric vision learners and generators, and cycle-consistent
text-guided image editing, sit adjacent to the representation and
prompt-revision tools used above
\citep{xu2025next,xu2026embedding,xu2023cyclenet}.

\section{Other generations of the same cards}
\label{sec:other-cells}
\label{app:other-cells}
\label{app:transfer-pilots}
\label{app:transfer-mist43}
\label{app:h3-transfer}
\label{app:v5}

The MIST cards name the same objects and operations throughout.
The experiments do not. Table~\ref{tab:transfer-cells} lists the
generator, the duration, the judge, and the claim for each reported
cell. Headline MIST is Wan TI2V-5B, three 5-second shots, and the
locked 122B judge. Official WRBench and MBench-T are Wan A14B, 10
seconds, and the released scorers. MiniMax-H3 pilots are packed 15s
Internal takes. A number from one row is not evidence for another row.
The strip in Fig.~\ref{fig:method} is the Internal MiniMax-H3 packed
take of \texttt{R0-03}: not the Wan TI2V-5B $n{=}43$ stitch, not the
equal-budget comparison in Table~\ref{tab:repair}, and not Official
$T$/$U$.

\begin{table}[t]
\caption{\textbf{Same cards, different experiments.}
Do not add across rows. Continuity is not reveal.}
\label{tab:transfer-cells}
\centering
\small
\setlength{\tabcolsep}{4pt}
\begin{tabular}{@{}p{0.22\linewidth}p{0.18\linewidth}p{0.10\linewidth}p{0.16\linewidth}p{0.26\linewidth}@{}}
\toprule
\textbf{Cell} & \textbf{Generator} & \textbf{$n$} & \textbf{Judge} &
\textbf{What it is} \\
\midrule
MIST kept reveal &
TI2V-5B $3{\times}5$s & 43 & 122B &
$0.147\!\rightarrow\!0.492$; Internal diagnosis \\
MIST best-of-five &
same & 43 & 122B &
$0.763$; oracle, answer in the prompt \\
Official $T$ &
A14B, 10s & 40 & official &
$0.452\!\rightarrow\!0.570$; leave--return \\
Official $U$ &
A14B, 10s & 30 & official &
$0.292\!\rightarrow\!0.446$; mostly keep baseline \\
H3 $n{=}6$ reveal &
H3, 15s packed & 6 & Internal 8B &
$0.435\!\rightarrow\!1.000$; a six-case pilot \\
H3-43 continuity &
H3, 15s packed & 43 & MAE $\ge 22$ &
$0.605\!\rightarrow\!0.721$; not memory \\
H3-43 122B reveal &
H3, 15s packed & 43 & 122B &
$0.387\!\rightarrow\!0.729$; never-open $3\!\rightarrow\!6$ \\
Figs.~\ref{fig:wording},~\ref{fig:h3-m3-gallery} &
H3, 15s packed & 6 stills & eye &
still pack, not seed 61001 \\
\bottomrule
\end{tabular}
\end{table}

\thinparagraph{A packed 15s take fails differently from the stitch.}
The headline $41/43$ never-open rate is a Wan TI2V-5B stitch of three
5-second shots. The same 43 cards can be packed into one MiniMax-H3
15s take: T2VA, short edge 768, seed 61001. The raw prompt does not
include the last-frame state sentence; the M3 arm adds that sentence.
Prompt expansion is not used in this 43-case run.
Table~\ref{tab:transfer-mist43-reveal} applies the locked 122B reveal
judge, with the same prompt and positions as Table~\ref{tab:diagnosis}.
The raw take opens in 40 of 43 cases. Its failure is a wrong exposed
state (24/43), not a closed lid. The never-open cell is therefore a
property of the stitch, not a property of every generator.
These are Internal packed cells, one clip per arm, with no keep-best.
They are not Official $T$ or $U$, and they are not the population of
Table~\ref{tab:diagnosis}.

\begin{table}[t]
\caption{\textbf{Internal 43-case H3 reveal.} Locked 122B, same
positions as MIST. Inside is the match rate of the exposed state.
Not the stitch population, and not Official.}
\label{tab:transfer-mist43-reveal}
\centering
\small
\begin{tabular}{@{}lcccccc@{}}
\toprule
\textbf{Arm} & \textbf{Reveal} & \textbf{Inside} & \textbf{Never-open} &
\textbf{Empty} & \textbf{Wrong} & \textbf{Ok} \\
\midrule
H3 raw & 0.387 & 0.256 & 3 & 6 & 24 & 10 \\
H3 + M3 & 0.729 & 0.674 & 6 & 1 & 8 & 28 \\
\bottomrule
\end{tabular}
\end{table}

Continuity asks only whether the 15s take hard-cuts. A frame pair
fails when the gray-frame mean absolute error is at least 22.
Table~\ref{tab:transfer-mist43-cont} is not a memory score, and
$0.605\rightarrow 0.721$ is not reveal and not Official $T$ or $U$.
M3 raises the pass rate and is not monotone: 12 cases fail raw and
pass M3, while 7 pass raw and fail M3. Both arms pass on 19 of 43
(Table~\ref{tab:transfer-mist43-fam}).

\begin{table}[t]
\caption{\textbf{Internal 43-case H3 continuity.} Adjacent-frame
harness. A hard cut is gray MAE $\ge 22$. Not a memory score.}
\label{tab:transfer-mist43-cont}
\centering
\small
\begin{tabular}{@{}lccc@{}}
\toprule
\textbf{Arm} & \textbf{$n$} & \textbf{Harness pass} & \textbf{Mean max MAE} \\
\midrule
H3 raw & 43 & 0.605 & 19.8 \\
H3 + M3 & 43 & 0.721 & 17.93 \\
\bottomrule
\end{tabular}
\end{table}

\begin{table}[t]
\caption{\textbf{Internal 43-case continuity by family.}
Adjacent-frame harness only. Not a memory score, and not
Table~\ref{tab:transfer-subset6}.}
\label{tab:transfer-mist43-fam}
\centering
\small
\begin{tabular}{@{}lcccc@{}}
\toprule
\textbf{Family} & \textbf{$n$} & \textbf{Raw pass} & \textbf{M3 pass} & \textbf{Both pass} \\
\midrule
R0 & 16 & 0.625 & 0.688 & 9 \\
R1 & 11 & 0.545 & 0.818 & 5 \\
R2 & 16 & 0.625 & 0.688 & 5 \\
All & 43 & 0.605 & 0.721 & 19 \\
\bottomrule
\end{tabular}
\end{table}

\thinparagraph{A six-case pilot is not a scale result.}
Table~\ref{tab:transfer-subset6} uses an Internal 8B judge on six
cards. H3 and Lingbot are separate generation pipelines. Neither uses
an official memory protocol. The arms are mixed rather than monotone:
H3 prompt expansion lowers the middle-shot score while raising reveal,
stacking M3 and expansion is worse than either arm alone, and Lingbot
stays near the floor. The diagnose-and-rewrite workflow can be run as
a packed 15s prompt on MiniMax-H3. That does not support a claim about
scale or about an official protocol. The planned 12-case H3 expansion
was paused before scoring. Lingbot M3 arms were not run.
We also do not fill in cells that were not run: a VideoRepair or
Video-T1 comparison at $K{=}5$ (the released Video-T1 code drives
Pyramid-Flow, not Wan, and was not ported), or human labels on the 43
last frames.

\begin{table}[t]
\caption{\textbf{Six-case Internal transfer pilots.}
Small $n$, a changed generator, and an Internal judge. Not a
model-scale result and not Official $T$ or $U$.}
\label{tab:transfer-subset6}
\centering
\small
\begin{tabular}{@{}lcccc@{}}
\toprule
\textbf{Pipeline / arm} & \textbf{Overall} & \textbf{Establish} &
\textbf{Middle shot} & \textbf{Reveal} \\
\midrule
H3 raw & 0.500 & 0.682 & 0.785 & 0.435 \\
H3 + prompt expansion & 0.937 & 0.842 & 0.503 & 1.000 \\
H3 + M3 & 0.937 & 0.747 & 0.685 & 1.000 \\
H3 + M3 + expansion & 0.855 & 0.829 & 0.665 & 0.879 \\
Lingbot raw & 0.035 & 0.008 & 0.025 & 0.042 \\
Lingbot + expansion & 0.311 & 0.790 & 0.569 & 0.190 \\
\bottomrule
\end{tabular}
\end{table}

\begin{figure}[t]
\centering
\includegraphics[width=\linewidth]{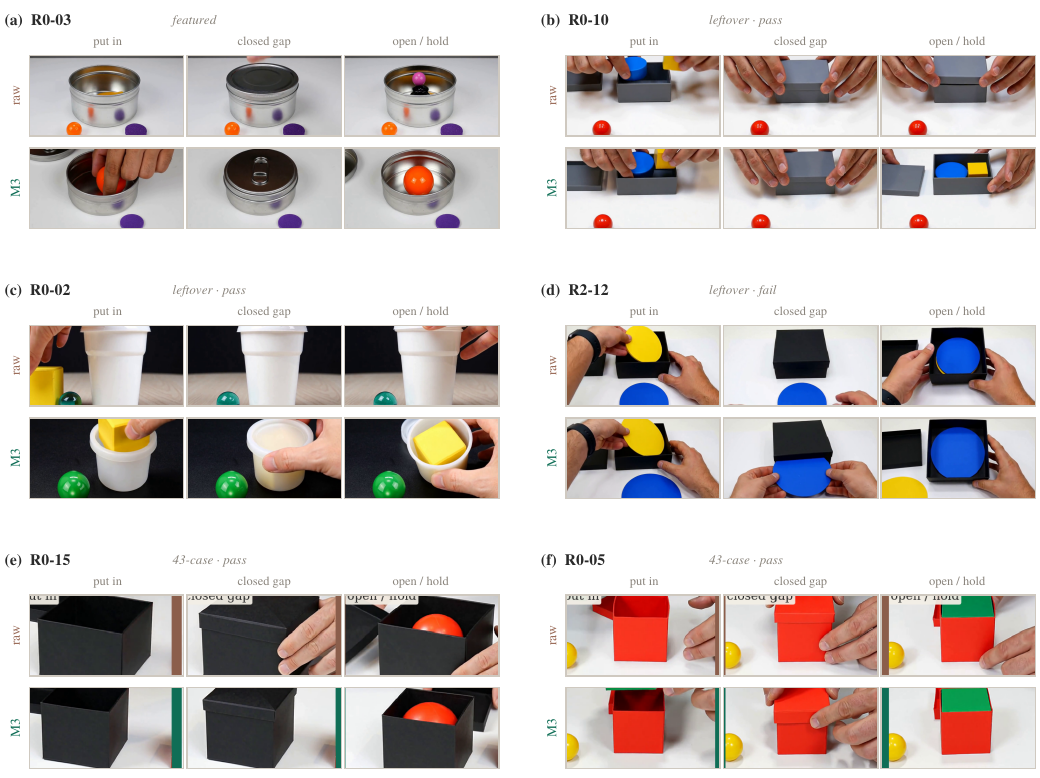}
\caption{\textbf{H3 raw versus M3 on the same cards.}
Each panel is put-in $\rightarrow$ closed gap $\rightarrow$ open/hold;
top row raw, bottom row M3.
(a)~\texttt{R0-03} (also Fig.~\ref{fig:method}).
(b--c)~other leftover high-pass takes.
(d)~leftover fail: both arms break continuity and show the wrong object.
(e--f)~seed-61001 43-case pairs, not the leftover pack.
Internal qualitative evidence only---not Official $T$/$U$, not
Table~\ref{tab:transfer-subset6}, and not reveal scores.}
\label{fig:h3-m3-gallery}
\end{figure}

Figure~\ref{fig:h3-m3-gallery} keeps the stills that this distinction
depends on. Panels (a--d) are one still pack. Panels (e--f) are the
seed-61001 run. The packs are not the same files: \texttt{R0-02}
passes the continuity harness in the still pack and fails it in the
43-case run. Panel (a) is the \texttt{R0-03} take in
Fig.~\ref{fig:wording}. Panel (d), \texttt{R2-12}, fails the harness
in both arms, and the last frame shows the wrong object.

\thinparagraph{Internal rescores and runs that produced no clips.}
Table~\ref{tab:app-external} rescores some of the same videos with a
different judge. Official $T$ and Official $U$ stay the rows already
reported. The Internal reveal on WRBench, $0.372\rightarrow 0.668$,
and the Internal unique score on a 12-case MBench-T slice,
$0.230\rightarrow 0.426$, use a different scale. They are not added to
$T$ or $U$. A scored intermediate pack at $0.463$ is not an official
cell and is not a claim of this paper. Official $U$ remains $0.446$.

\begin{table}[t]
\caption{\textbf{Official and Internal scores on the same suite names.}
Do not add across rows. WRBench $T$ scores a missing return as zero.
The visual-quality row uses a different clip on 24 of 40 items.}
\label{tab:app-external}
\centering
\small
\begin{tabular}{@{}llcccc@{}}
\toprule
\textbf{Suite} & \textbf{Cell} & \textbf{Base} & \textbf{M3} & $\boldsymbol{\Delta}$ & \textbf{Track} \\
\midrule
WRBench/40 & Official $T$ & 0.452 & 0.570 & +0.118 & Official \\
WRBench/40 & Official vis.\ (other clip) & 0.648 & 0.755 & +0.107 & Official \\
MBench-T/30 & Official $U$ & 0.292 & 0.446 & +0.154 & Official \\
WRBench/40 & Internal reveal & 0.372 & 0.668 & +0.296 & Internal \\
MBench-T/12 & Internal unique & 0.230 & 0.426 & +0.196 & Internal \\
\bottomrule
\end{tabular}
\end{table}

One further protocol was attempted and did not yield a score. It
paired TI2V-5B aftermath stills with an A14B image-to-video stitch of
161 frames, seed 10042, on 12 items, instead of frozen T2V-A14B.
The run failed before producing clips (FSDP \texttt{no\_sync}; 0/12),
so it has no official score. Adapted MemoBench and SANA-WM smoke tests
are excluded for the same reason: their generation or metrics are not
the protocol in Table~\ref{tab:protocol-cards}.

\section{Held-out last-frame judge}
\label{app:judge}
\label{app:who}

The main 122B judge is unchanged. Qwen3-VL-8B-Instruct rereads the
same locked last frames as Table~\ref{tab:who} and answers YES or NO
to $s_T^\star$. Official $T$ and $U$ are not replaced.
The main-text last-frame rate $0.023$ is a strict YES rate. Mean
inside in Table~\ref{tab:diagnosis} is $0.047$.
Agreement with the 122B inside check is high on the baseline because
almost every baseline frame is negative. Agreement on the action bit
of rewritten frames is much lower, and agreement with official WRBench
and MBench gates is lower still
(Table~\ref{tab:app-judge}). An 8B YES is not an official $D5$ or
$D6$ decision, and it is not an MBench state score. Those rubrics ask
different questions.

\begin{table}[H]
\caption{\textbf{Held-out 8B against the locked 122B judge and the
official gates.} MIST inside agreement is high under class imbalance.
An 8B YES is not official $D5$, $D6$, or MBench state.}
\label{tab:app-judge}
\centering
\small
\begin{tabular}{@{}lccc@{}}
\toprule
\textbf{Set} & \textbf{8B YES} & \textbf{Reference rate} & \textbf{Agree} \\
\midrule
MIST last, base (vs 122B inside) & 0.09 & 0.023 & 0.93 \\
MIST last, M3 (vs 122B inside) & 0.56 & 0.442 & 0.88 \\
MIST last, base (vs 122B act) & 0.09 & 0.070 & 0.84 \\
MIST last, M3 (vs 122B act) & 0.56 & 0.535 & 0.51 \\
WRBench unique last (vs D5 gate) & 0.33 & 0.78 & 0.50 \\
WRBench unique last (vs D6 gate) & 0.33 & 0.78 & 0.35 \\
MBench winner last (vs state${>}0.5$) & 0.17 & 0.47 & 0.43 \\
\bottomrule
\end{tabular}
\end{table}

\end{document}